\documentclass[a4paper, 12pt, twocolumn]{article} % For LaTeX2e

\usepackage{graphicx}
\usepackage{mathrsfs}
\usepackage{subcaption}
\usepackage{array}
\usepackage{amsfonts}
\usepackage{amsmath}
\usepackage{hyperref}
\usepackage{eurosym}
\usepackage{multirow}
\usepackage{url}
\usepackage{xcolor}
\usepackage{pbox}
\usepackage[top=3cm, bottom=2.5cm, left=2cm, right=2cm]{geometry}

\usepackage[T1]{fontenc}
\usepackage[sfdefault]{AlegreyaSans}

\title{ShaResNet: reducing residual network parameter number by sharing weights}
\author{Alexandre Boulch \\ \textit{ONERA, The French Aerospace Lab, F-91761 Palaiseau, France}}
\date{}

\begin{document}
    \maketitle

    {
    \color{red}
    \bf Note This paper is currently under consideration at Pattern Recognition Letters.
    }

    \noindent\makebox[\linewidth]{\rule{\linewidth}{1pt}}

    \begin{abstract}
    Deep Residual Networks have reached the state of the art in many image processing tasks such image classification. However, the cost for a gain in accuracy in terms of depth and memory is prohibitive as it requires a higher number of residual blocks, up to double the initial value. To tackle this problem, we propose in this paper a way to reduce the redundant information of the networks. We share the weights of convolutional layers between residual blocks operating at the same spatial scale. The signal flows multiple times in the same convolutional layer. The resulting architecture, called ShaResNet, contains block specific layers and shared layers. These ShaResNet are trained exactly in the same fashion as the commonly used residual networks. We show, on the one hand, that they are almost as efficient as their sequential counterparts while involving less parameters, and on the other hand that they are more efficient than a residual network with the same number of parameters. For example, a 152-layer-deep residual network can be reduced to 106 convolutional layers, i.e. a parameter gain of 39\%, while loosing less than 0.2\% accuracy on ImageNet.

    \end{abstract}

    \noindent\makebox[\linewidth]{\rule{\linewidth}{1pt}}

\begin{figure}
    \centering
    \includegraphics[angle=-90,width=0.91\linewidth]{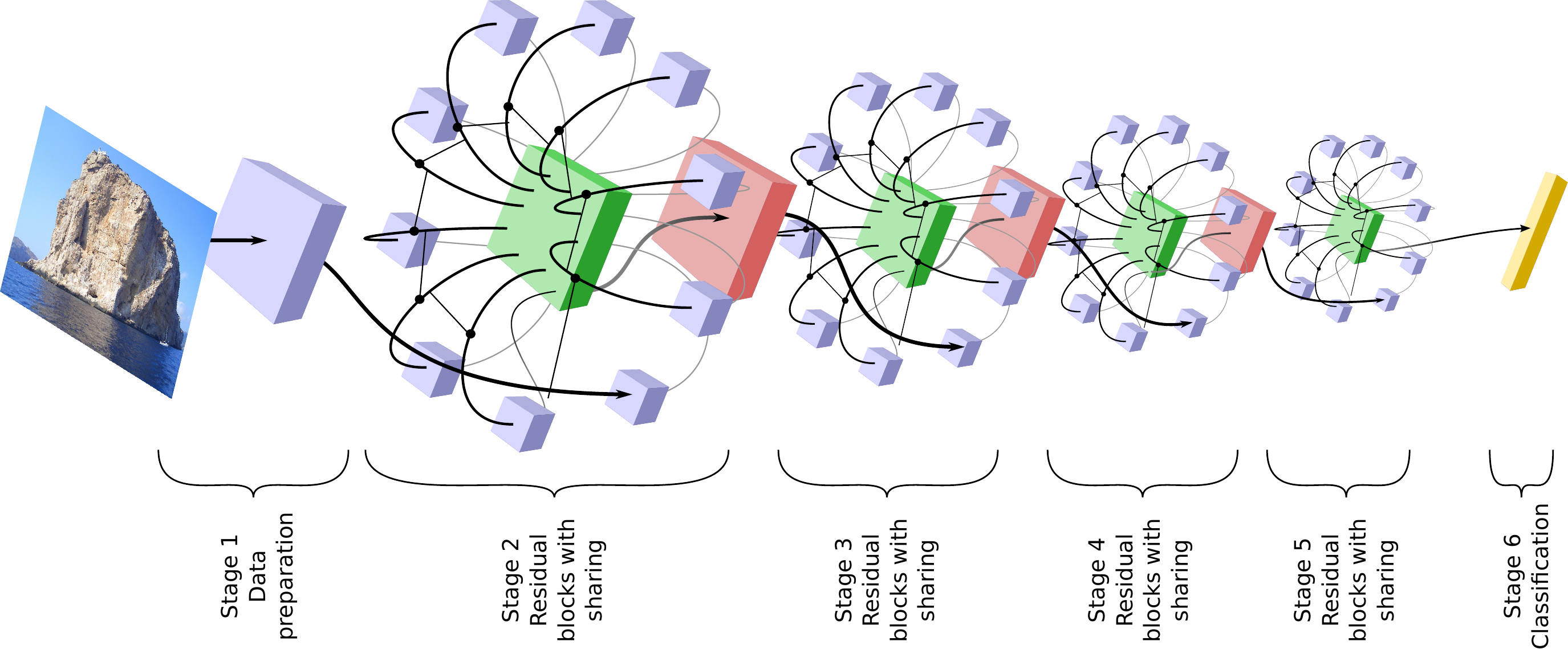}

    \textit{Color code: green convolutions are shared between several residual blocks, red blocks are spatial dimensionality reductions (covolution or pooling), blue ones are the other convolutions and the yellow block is the classifier, average pooling or fully connected.}

    \caption{3D representation of a residual network with shared convolution.}
    \label{fig:share1}
\end{figure}

%%%%%%%%%%%%%%%%%%%%%%%%%%%%%%%%%%%%%%%%%%%%%%%%%%%%%%%%%%%%%%%%%%%%%%%%%%%%
%%%%%%%%%%%%%%%%%%%%%%%%%%%%%%%%%%%%%%%%%%%%%%%%%%%%%%%%%%%%%%%%%%%%%%%%%%%%
% #### ##    ## ######## ########   #######
%  ##  ###   ##    ##    ##     ## ##     ##
%  ##  ####  ##    ##    ##     ## ##     ##
%  ##  ## ## ##    ##    ########  ##     ##
%  ##  ##  ####    ##    ##   ##   ##     ##
%  ##  ##   ###    ##    ##    ##  ##     ##
% #### ##    ##    ##    ##     ##  #######
%%%%%%%%%%%%%%%%%%%%%%%%%%%%%%%%%%%%%%%%%%%%%%%%%%%%%%%%%%%%%%%%%%%%%%%%%%%%
%%%%%%%%%%%%%%%%%%%%%%%%%%%%%%%%%%%%%%%%%%%%%%%%%%%%%%%%%%%%%%%%%%%%%%%%%%%%

    \section{Introduction}

    Convolutional Neural Networks (CNNs) are now widely used for image processing tasks from classification~\cite{lecun1998gradient} and object detection~\cite{girshick14CVPR,renNIPS15fasterrcnn} to semantic segmentation~\cite{badrinarayanan2015segnet,audebert2016semantic}.
    Their utilisation even generalizes to other fields where data can be represented as tensors like in point cloud processing~\cite{boulch2016deep} or 3D shape style identification~\cite{10.1111:cgf.12977}.
    Today's network architectures still carry a strong inheritance of the CNN early stage designs.
    They are based on stacking convolutional, activation and dimensionality reduction layers.
    Over the past years, the progress in image processing tasks went together with a gradual increase in the number of layers, from AlexNet~\cite{krizhevsky2012imagenet} to Residual networks~\cite{he2016deep} (ResNets) that may contain up to hundreds of convolutions.

    Practical use of such networks may be challenging when using low memory system, such as autonomous vehicles, both for optimization and inference.
    Moreoever, from a biological point of view, a higher number of stacked layers leads to networks further from the original underlying idea of neural networks: biological brain mimicry.
    According to the current knowledge of the brain, cerebral cortex is composed of a low number of layers where the neurons are highly connected.
    Moreover the signal is also allowed to recursively go through the same neurons.
    In that sense recurrent neural networks are much closer to the brain structure but more difficult to optimize~\cite{Bengio_learninglong-term,hochreiter1997long}.

    Looking more closely at the repetition of residual blocks in ResNets, it could somehow be interpreted as an unwrapped recurrent neural networks.
    This constatation raises questions such as \textit{"how similar are the weights of the blocks ?"}, \textit{"do the same parts of the blocks operate similar operations ?"} and in the later case \textit{"is it possible to reduce the parameter number of a residual network ?"}.
    Driven by these observations and questions, we present a new network architecture based on residual networks where part of the convolutions share weights, called \textit{ShaResNets}.
    It results in a great decrease of the number of network parameters, from $25\%$ to $45\%$ depending on the size of the original architecture.
    Our networks also present a better ratio performances over parameter number while downgrading the absolute performance by less than $1\%$.

    The paper is organized as follow: section~\ref{sec:related} presents the related work on convolutional neural networks (CNNs) ; the ShaResNets are presented in section~\ref{sec:sharesnet} and finally, in section~\ref{sec:expe}, we expose our experimentations on classification datasets CIFAR 10 and 100 and ILSVRC Imagenet.

%%%%%%%%%%%%%%%%%%%%%%%%%%%%%%%%%%%%%%%%%%%%%%%%%%%%%%%%%%%%%%%%%%%%%%%%%%%%
%%%%%%%%%%%%%%%%%%%%%%%%%%%%%%%%%%%%%%%%%%%%%%%%%%%%%%%%%%%%%%%%%%%%%%%%%%%%
% ########  ######## ##          ###    ######## ######## ########
% ##     ## ##       ##         ## ##      ##    ##       ##     ##
% ##     ## ##       ##        ##   ##     ##    ##       ##     ##
% ########  ######   ##       ##     ##    ##    ######   ##     ##
% ##   ##   ##       ##       #########    ##    ##       ##     ##
% ##    ##  ##       ##       ##     ##    ##    ##       ##     ##
% ##     ## ######## ######## ##     ##    ##    ######## ########
%%%%%%%%%%%%%%%%%%%%%%%%%%%%%%%%%%%%%%%%%%%%%%%%%%%%%%%%%%%%%%%%%%%%%%%%%%%%
%%%%%%%%%%%%%%%%%%%%%%%%%%%%%%%%%%%%%%%%%%%%%%%%%%%%%%%%%%%%%%%%%%%%%%%%%%%%

    \section{Related work}
    \label{sec:related}

    CNNs were introduced in~\cite{lecun1989backpropagation} for hand written digits recognition.
    They became over the past years one of the most enthusiastic field of deep learning~\cite{lecun2015deep}.
    The CNNs are usually built using a common framework.
    They contains many convolutional layers and operate a gradual spatial dimension reduction using convolutional strides or pooling layers~\cite{ciregan2012multi,jarrett2009best}.
    This structure naturally integrates low/mid/high level features along with a dimension compression before ending with a classifier, commonly a perceptron~\cite{rosenblatt1957perceptron}, multi-layered or not, i.e. one or more fully connected layers.

    Looking at the evolution CNNs, the depth appears to be a key feature.
    On challenging image processing tasks, an increase of performance is often related to a deeper network.
    As an example, AlexNet~\cite{krizhevsky2012imagenet} has 5 convolutions while VGG16 and VGG19~\cite{simonyan2014very} have respectively 16 and 19 convolutional layers and more recently, in~\cite{huang2016deep}, the authors train a 1200 layer deep network.

    Variations in the LeNet structure have also been used to improve convergence.
    In a Network in Network (NiN)~\cite{lin2013network}, convolutions are mapped with a multilayer perceptron (1x1 convolutions), which prevent overfitting and improved accuracy on datasets such as CIFAR~\cite{krizhevsky2009learning}.
    GoogLeNet~\cite{szegedy2015going} introduced a multiscale approach using the inception module, composed of parallel convolutions with different kernel sizes.

    Optimizing such deep architectures can face practical problems such as overfitting or vanishing or exploding gradients.
    To overcome these issues, several solutions have been proposed such as enhance optimizers~\cite{sutskever2013importance}, dropout~\cite{srivastava2014dropout} applying a random reduction of the number of connection in fully connected layers or on convolutional layers~\cite{DBLP:journals/corr/ZagoruykoK16}, intelligent initialization strategies~\cite{glorot2010understanding} or training sub-networks with stochastic depth~\cite{huang2016deep}.

    Residual networks~\cite{he2016deep} achieved the state of the art in many recognition tasks including Imagenet~\cite{ILSVRC15} and COCO~\cite{lin2014microsoft}.
    They proved to be easier to optimize.
    One of the particularities of these networks is to be very deep, up to hundreds of residual layers.
    More recently, the authors of~\cite{DBLP:journals/corr/ZagoruykoK16} introduced wide residual networks which reduce the depth compared to usual resnets by using wider convolutional blocks.

    Reducing network size has been the object of several works.
    More compact layers are also used, like the replacement of the fully connected layers by average pooling~\cite{szegedy2015going, he2016deep}.
    In~\cite{DBLP:journals/corr/RastegariORF16,DBLP:journals/corr/CourbariauxB16}, the weights are constrained be binary, reducing considerably the memory consumption.
    \cite{pratt1989comparing} prune weights in pretrained networks, modfiying the structure, to create a lighter network.
    \cite{han2015deep} remove redundant connections and allow weight sharing.
    Similarly, in this work, we use weight sharing to reduce the network size, except that we enforce a predefined sharing structure at training time.
    We do not need post processing of the network.

%%%%%%%%%%%%%%%%%%%%%%%%%%%%%%%%%%%%%%%%%%%%%%%%%%%%%%%%%%%%%%%%%%%%%%%%%%%%
%%%%%%%%%%%%%%%%%%%%%%%%%%%%%%%%%%%%%%%%%%%%%%%%%%%%%%%%%%%%%%%%%%%%%%%%%%%%
% ##    ## ######## ########
% ###   ## ##          ##
% ####  ## ##          ##
% ## ## ## ######      ##
% ##  #### ##          ##
% ##   ### ##          ##
% ##    ## ########    ##
%%%%%%%%%%%%%%%%%%%%%%%%%%%%%%%%%%%%%%%%%%%%%%%%%%%%%%%%%%%%%%%%%%%%%%%%%%%%
%%%%%%%%%%%%%%%%%%%%%%%%%%%%%%%%%%%%%%%%%%%%%%%%%%%%%%%%%%%%%%%%%%%%%%%%%%%%

    \begin{figure*}[!ht]
        \centering
        \includegraphics[width=\linewidth]{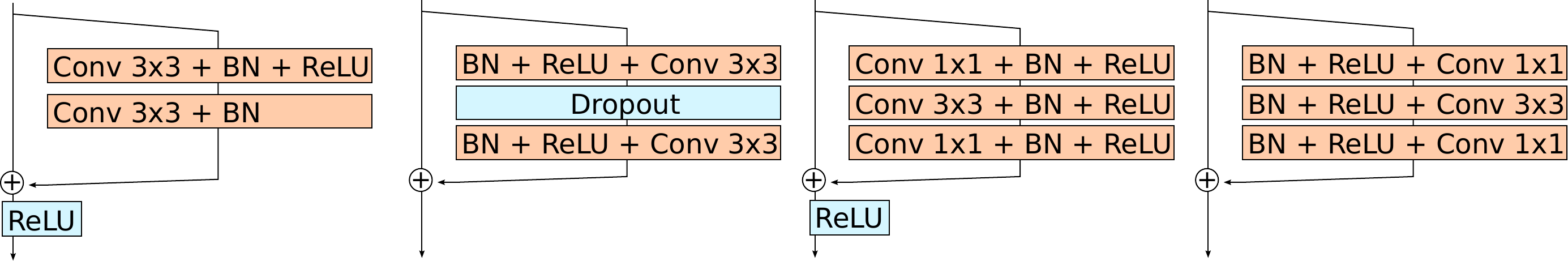}
        \textit{From left to right:
        Basic Residual Block (ImageNet),
        Basic Wide Residual Block with optional dropout (CIFAR),
        Bottleneck Residual Block (ImageNet) and
        Bottleneck Residual Block (CIFAR).}
        \caption{Residual blocks used for evaluation.}
        \label{fig:res_block}
    \end{figure*}

    \section{Sharing Residual Networks}
    \label{sec:sharesnet}

    ShaResNets are based on residual networks architectures in which we force the residual blocks in the same stage, i.e. between two spatial dimension reduction, to share the weights of one convolution.
    In this section, we first present the residual architectures we based our work on, and then, detail the sharing process.

    \subsection{Residual networks. }
    The residual networks basic~\cite{he2016deep} or wide~\cite{DBLP:journals/corr/ZagoruykoK16} are a sequential stack of residual blocks, with several convolution layers bypassed by parallel branch.
    The output of block $k$, $\mathbf{x}_{k+1}$ can be represented as:
    \begin{equation}
        \mathbf{x}_{k+1} = \mathbf{x}_{k} + \mathcal{F}(\mathbf{x}_{k}, W_{k})
        \label{eq:residual}
    \end{equation}
    where $\mathbf{x}_{k}$ is the input (output of block $k-1$), $\mathcal{F}$ is the residual function and $W_k$ are the parameters of the residual unit.
    Among them two types of residual blocks are used in this paper.
    \begin{itemize}
        \item \textit{basic} composed of two consecutive 3x3 convolutions.
        \item \textit{bottleneck} composed of one 3x3 convolution surrounded by two 1x1 convolutions for reducing and then expanding the dimensionality.
    \end{itemize}
    A common element of the convolutional structure in the residual blocks is at least a 3x3 convolution.
    This convolution allows a neighborhood connection so that the final decision is taken using neighborhood relations between pixels and not only independent pixel values.

    Figure~\ref{fig:res_block} describes the blocks composing the networks presented in this paper.
    We used two implementations, depending on the datasets (CIFAR 10-100 and ImageNet) to fit the original network structure we will compare to (section~\ref{sec:expe}).
    They differ in the position of batch normalization and ReLU, before convolutions for CIFAR datasets and after for ImageNet.

    \subsection{Sharing weights}

    % \begin{figure}
    %     \centering
    %     \begin{tabular}{cc}
    %         \includegraphics[width=0.48\linewidth]{images/sharing_2}&
    %         \includegraphics[width=0.48\linewidth]{images/sharing_3}\\
    %         \pbox{0.47\linewidth}{ \textit{On the left, the original residual block (top: bottleneck, bottom: basic) and on the right, the block discrimination operated before sharing: block specific and spatial relations.}} &
    %          \pbox{0.47\linewidth}{ \textit{The spatial relations are shared between all the residual blocks of the same stage (between two spatial dimension reduction).}}\\
    %         (a) & (b)
    %     \end{tabular}
    %
    %     \caption{Block specific operations and spatial operations to shared residual block.}
    %     \label{fig:share_bottleneck}
    % \end{figure}

    \begin{figure}[!ht]
        \centering
        \begin{tabular}{c}
            \includegraphics[width=0.9\linewidth]{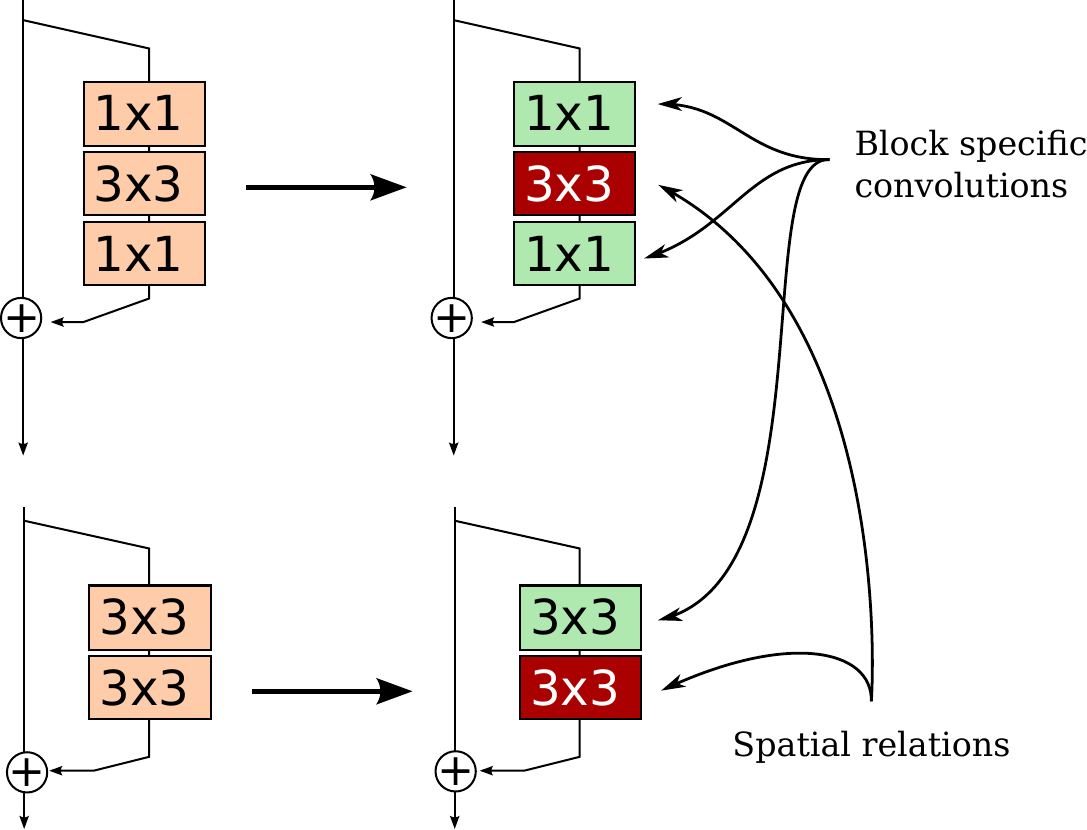}\\

            \pbox{0.9\linewidth}{ (a) \textit{On the left, the original residual block (top: bottleneck, bottom: basic) and on the right, the block discrimination operated before sharing: block specific and spatial relations.}} \\
            \includegraphics[width=0.9\linewidth]{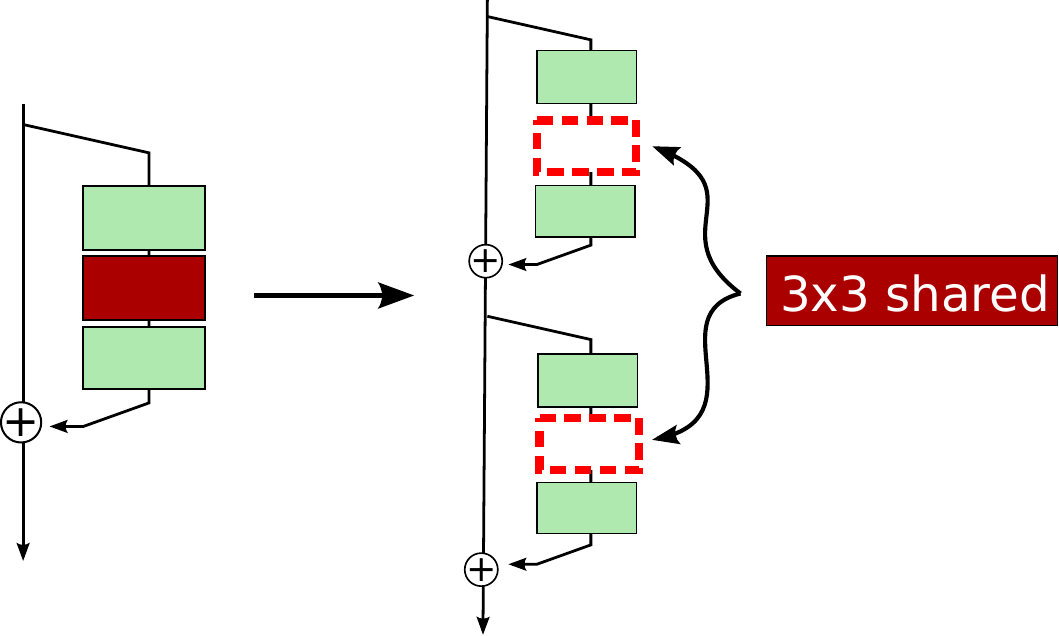}\\
             \pbox{0.9\linewidth}{ (b) \textit{The spatial relations are shared between all the residual blocks of the same stage (between two spatial dimension reduction).}}\\

        \end{tabular}

        \caption{Block specific operations and spatial operations to shared residual block.}
        \label{fig:share_bottleneck}
    \end{figure}

    The underlying idea of our approach is that it is possible to somehow distinguish two types of mechanisms in a residual block.
    The first, specific to the block, is the abstraction of the residual block.
    From block to block, it created higher level features.
    The second, redundant in the blocks of the same stage, is the spatial connection relations between neighboring tensor cells, between pixels.
    The equation~\ref{eq:residual} becomes:
    \begin{equation}
        \mathbf{x}_{s,k+1} = \mathbf{x}_{s,k} + \mathcal{F}(\mathbf{x}_{s,k}, W_{s,k}, W_{s})
        \label{eq:share}
    \end{equation}
    where $\mathbf{x}_{s,k}$ (resp. $\mathbf{x}_{s,k+1}$) is the input (resp. the output) of block $(k,s)$, $k$-th residual unit of stage $s$.
    $W_{k}$ of equation~\ref{eq:residual} is split into $W_{k,s}$, the parameters specific to the residual block and $W_{s}$ the parameters shared at the stage level.

    In ResNets, the spatial information is taken into account in the 3x3 convolutions and in the layers with dimension reduction (convolution with stride or pooling layers).
    We first look at the bottleneck block, composed of 3 convolutions (1x1, 3x3 and 1x1).
    The fictive separation between spatial connection and specific operations is easy as the 1x1 convolution do not connect neighboring cells.
    This is the top line of figure~\ref{fig:share_bottleneck}(a).
    Then, for all spatial connections in the same stage, i.e. for all the blocks between two pooling layers (or convolution with stride), we share the weights.
    By using a unique 3x3 convolution, we consider that all the spatial connections of a given stage can be explained by a common set of kernels (figure~\ref{fig:share_bottleneck}(b)).

    We adapt the approach to the basic residual block.
    As it is composed of two 3x3 convolutions, extracting the spatial component is not possible.
    Still, we adopt a similar approach, the first convolution is considered as specific and the second is shared with the blocks of the same stage (figure~\ref{fig:share_bottleneck}, bottom line).

    In the two cases, we obtain a similar global architecture represented in figure~\ref{fig:share1}.
    For each stage, the green convolution is common to all block while the specific items are blue (the number of specific items depends on the architecture choice).
    The red blocks are the dimensionality reduction layers and the yellow one would be either a multi-layer perceptron or an average pooling.

    \begin{table*}[!ht]
        \centering
        \begin{tabular}{|c|c|cc|c|cc|}
            \hline
            Dataset& Model & \multicolumn{2}{c|}{Parameter number} & Parameter & \multicolumn{2}{c|}{Convolution nbr.}\\
            & & Orignal & ShaResNet & decrease  & Orignal & ShaResNet\\
            \hline
            \multirow{2}{*}{CIFAR 10} &ResNet-164  & $1.70$ M & $0.93$ M & $45\%$ & 164 & 113\\
            &WRN-40-4    & $8.95$ M & $5.85$ M & $35\%$ & 40 & 25\\
            \hline
            CIFAR 100& WRN-28-10   & $36.54$ M & $26.86$ M & $26\%$ & 28 & 19\\
            \hline

            \multirow{4}{*}{IMAGENET}&ResNet-34   & $21.8$ M & $13.6$ M & $37\%$ & 34 & 20\\
            &ResNet-50   & $25.6$ M & $20.5$ M & $20\%$ & 50 & 38 \\
            &ResNet-101  & $44.5$ M & $29.4$ M & $33\%$ & 101 & 72 \\
            &ResNet-152  & $60.2$ M & $36.8$ M & $39\%$ & 152 & 106\\
            \hline
        \end{tabular}
        \caption{Number of parameters of the networks.}
        \label{t_param}
    \end{table*}

    \subsection{Gradient propagation}

    The gradient propagation in the block specific convolutions is similar to the usual stochastic gradient descent with momentum.
    The corresponding update rule for convolution $k$ of stage $s$ is:
     \begin{align}
         v_{s,k} &= \gamma v_{s,k} + \alpha \nabla_{W_{s,k}} J(W_{s,k})\\
         W_{s,k} &= W_{s,k} - v_{s,k}
     \end{align}
     where $\gamma$ is the momentum, $\alpha$ is the learning rate, $J(W_{s,k})$ is the objective function and $v$ is the velocity vector.

     In the case of shared convolution the gradients are accumulated before weight update.
     The update rule becomes:
    \begin{align}
      v_{s} &= \gamma v_{s} + \alpha \sum_{i \in S} \nabla_{W_{s}} J(W_{s})^{(i)}\\
      W_{s} &= W_{s} - v_{s}
    \end{align}
    where $i$ stands for the index of the block of stage $s$.

%%%%%%%%%%%%%%%%%%%%%%%%%%%%%%%%%%%%%%%%%%%%%%%%%%%%%%%%%%%%%%%%%%%%%%%%%%%%
%%%%%%%%%%%%%%%%%%%%%%%%%%%%%%%%%%%%%%%%%%%%%%%%%%%%%%%%%%%%%%%%%%%%%%%%%%%%
% ######## ##     ## ########  ########
% ##        ##   ##  ##     ## ##
% ##         ## ##   ##     ## ##
% ######      ###    ########  ######
% ##         ## ##   ##        ##
% ##        ##   ##  ##        ##
% ######## ##     ## ##        ########
%%%%%%%%%%%%%%%%%%%%%%%%%%%%%%%%%%%%%%%%%%%%%%%%%%%%%%%%%%%%%%%%%%%%%%%%%%%%
%%%%%%%%%%%%%%%%%%%%%%%%%%%%%%%%%%%%%%%%%%%%%%%%%%%%%%%%%%%%%%%%%%%%%%%%%%%%

    \section{Experimental results}
    \label{sec:expe}

    \subsection{Datasets and architectures}

    We experiment on three dataset, CIFAR 10, CIFAR 100 and ImageNet.
    We propose evaluations consisting into a comparison between the ShaResNets and their ResNet counterpart.

    CIFAR 10 and 100 are two datasets containing 50000 images for training and 10000 for test.
    In order to show that our sharing process can be generalized to different residual architectures, we use ResNets and Wide ResNets:
    \begin{itemize}
        \item The ResNets implementation (ResNet-164) with 164 convolutions is a good example of very deep network for CIFAR 10 dataset, based on basic residual blocks. (figure~\ref{fig:res_block} second column).
        \item Wide residual networks are not as deep as the previous but are composed of wider convolutions (more convolutional planes).
        We present results with two depth: 40 (WRN-40-4) for CIFAR 10 and 28 (WRN-28-10) for CIFAR 100. These are based on the wide residual block (figure~\ref{fig:res_block} first column). The dropout is only activated for CIFAR 100.
    \end{itemize}

    Imagenet is a much bigger dataset, with more than one million training images and 1000 classes.
    We experiment with residual networks of different depth: 34 (ResNet-34) with basic block (figure~\ref{fig:res_block} third column), 50, 101 and 152 (ResNet-50 and ResNet-152) with bottleneck block (figure~\ref{fig:res_block} last column).

    \subsection{Parameter number reduction}

    This section deals with the consequences of sharing convolutional weights on the network parameter number.
    Using one convolution for the spatial relations per stage instead of one per block reduce significantly the size of the network.
    Comparatively, the deeper the ResNet is the the bigger the gain is for its ShaResNet version.

    Table~\ref{t_param} shows the figures for the different architectures and datasets.
    As expected, the gain is substantial, from $20\%$ for ResNet-50 (ImageNet) to $45\%$ for ResNet-164 (CIFAR).
    The convolution number in the table expresses the number of independant convolutionnal layers, the shared convolutional layers are counted once.

    \subsection{Training}

\begin{figure}

        \centering
        \includegraphics[width=0.48\linewidth]{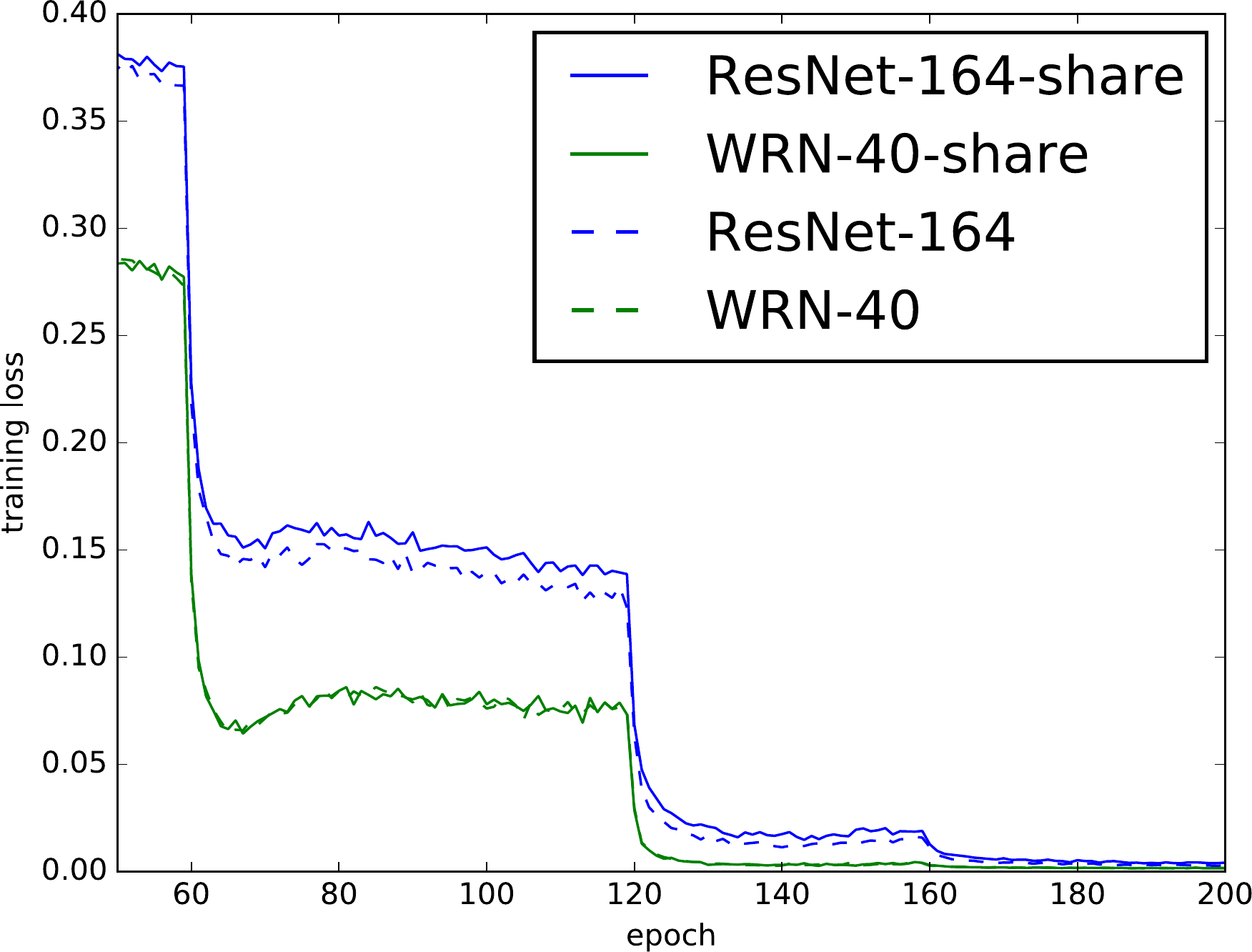}~~
        \includegraphics[width=0.48\linewidth]{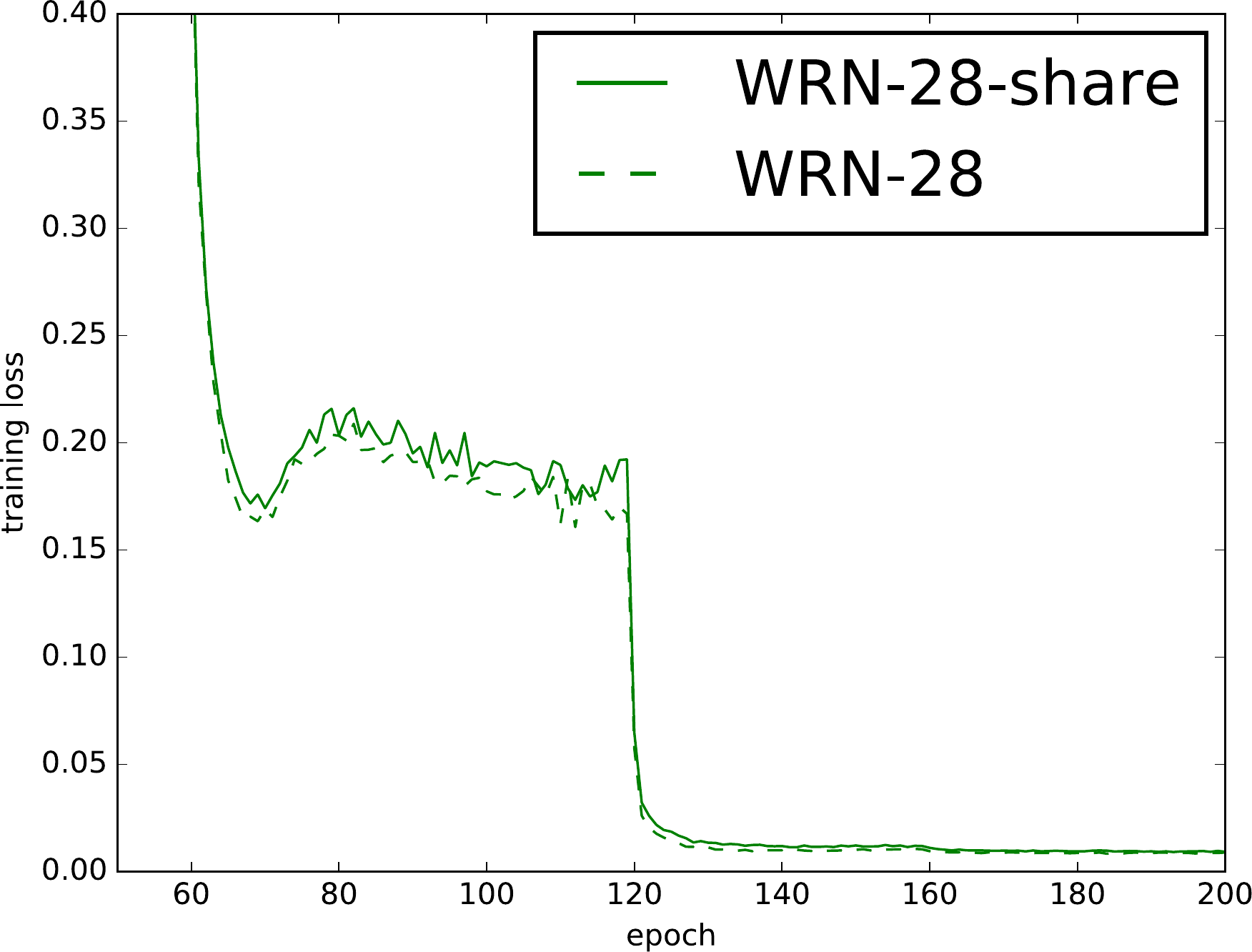}\\
        \textit{Training loss on CIFAR 10 (left) and CIFAR 100 (right).}

        \includegraphics[width=0.48\linewidth]{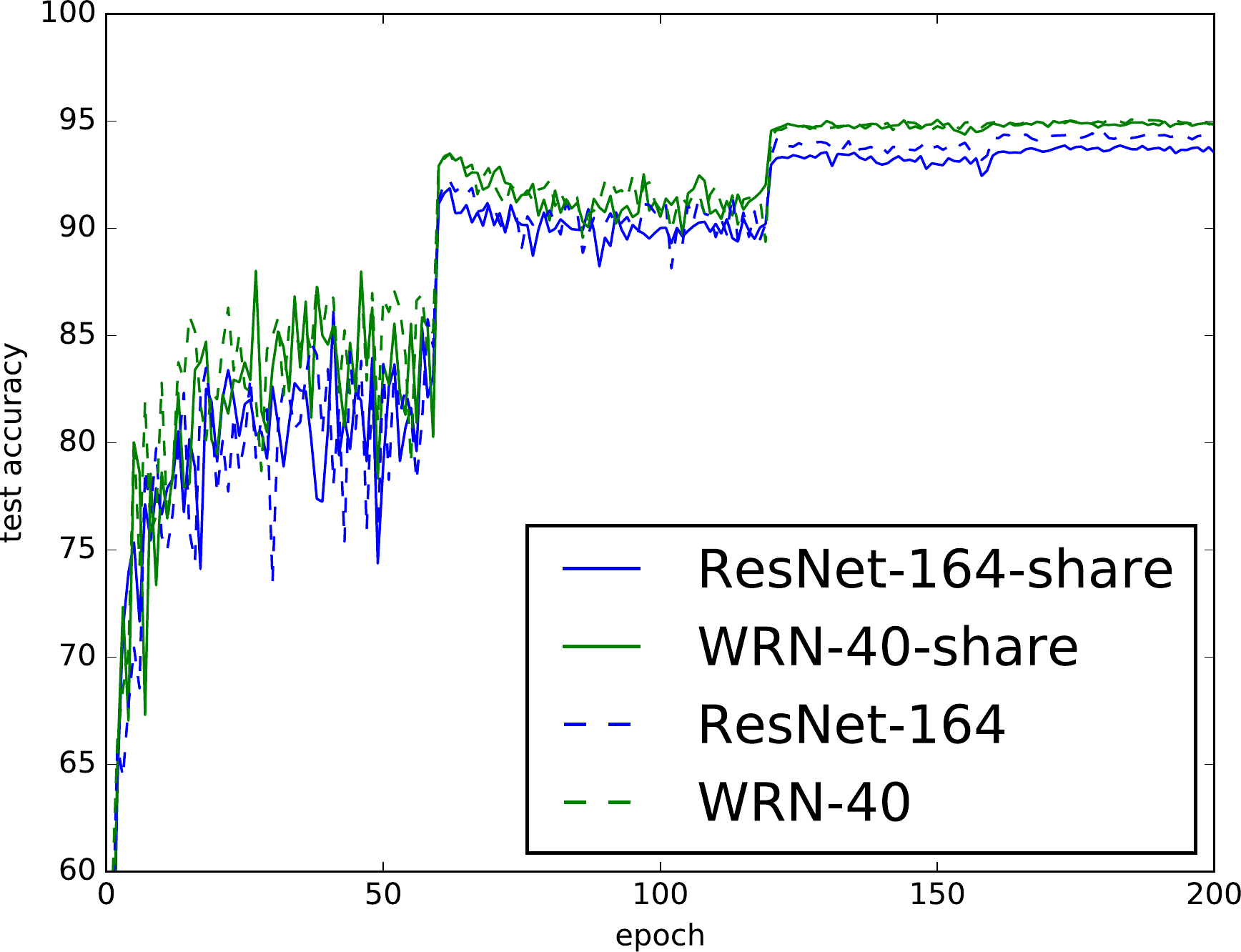}~~
        \includegraphics[width=0.48\linewidth]{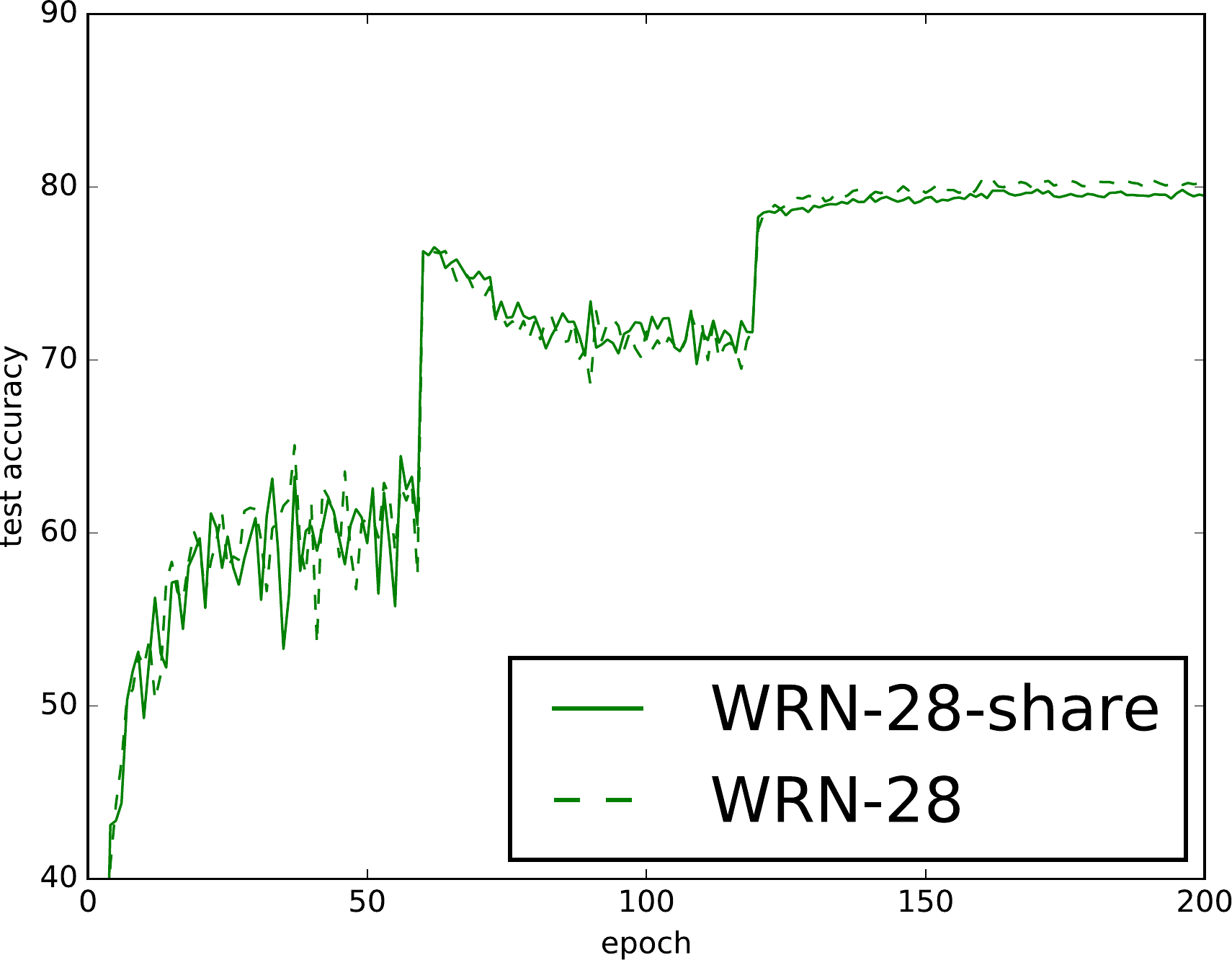}\\
        \textit{Test accuracy on CIFAR 10 (left) and CIFAR 100 (right).}

        \caption{Training loss and test accuracy on CIFAR}
        \label{fig:test_train_acc}
    \end{figure}

    ShaResNet are trained using the same training process as their non-shared counterpart.
    We used a stochastic gradient descent with momentum and step dropping learning rate policy.
    CIFAR models were trained with whitened data from PyLearn2~\cite{pylearn2_arxiv_2013} and we applied a random horizontal flip on the input image to simulate a larger training dataset and avoid over-fitting.
    For ImageNet, we adopted a similar approach, we apply on the input image a random crop, random contrast, lighting and color normalization as well as horizontal flip.
    According to our experiments, the behaviors of our networks are very similar to the original ones.
    Figure~\ref{fig:test_train_acc} presents the training loss and testing accuracy plots obtained on the CIFAR datasets.
    The gradient accumulation at shared convolutions does not induces instabilities at both training and testing time.

    \subsection{Accuracy}

    \begin{table*}[!ht]
        \centering
        \begin{tabular}{|c|c|ccc|ccc|}
            \hline
            Dataset& Model  & \multicolumn{3}{c|}{Error top 1 ($\%$)} & \multicolumn{3}{c|}{Error top 5 ($\%$)}\\
                 && Orig. & Share. & Diff. & Orig. & Share. & Diff.\\
             \hline
             \multirow{2}{*}{CIFAR 10} &ResNet-164 &  $94.54$ & $93.8$ & $0.74$ &&&\\
             &WRN-40-4   &  $95.83$ & $94.9$ & $0.93$ &&&\\
             \hline
             CIFAR 100 &WRN-28-10   & $80$ & $79.8$ & $0.2$  &&&\\
            \hline
            \multirow{4}{*}{IMAGENET}&ResNet-34&$26.73$& $28.25$&$0.52$ & $8.74$ & $9.42$ &$0.66$\\
            &ResNet-50   & $24.01$  & $24.61$ & $0.6$  & $7.02$ & $7.41$ &$0.39$\\
            &ResNet-101  & $22.44$  & $22.91$ & $0.47$ & $6.21$ & $6.55$ &$0.34$\\
            &ResNet-152  & $22.16$  & $22.23$ & $0.07$ & $6.16$ & $6.14$ &$-0.02$\\
            \hline
        \end{tabular}

        \caption{Accuracy on CIFAR and ImageNet. Error percentage (top 1 and top 5) and gap between original and shared version.}
        \label{tab:acc_datasets}
    \end{table*}

    Quantitative evaluation of the networks are presented on table~\ref{tab:acc_datasets}.
    On all these experiments, the top-1 decrease by less than $1\%$ when using the ShaResNet version of the algorithm.
    The gap between original and shared version is lower on ImageNet and CIFAR 100.
    This is to be related to the sizes of the networks.
    The CIFAR 10 networks are smaller than the others (less than $10$M parameters).
    To our understanding, smaller residual networks induces less redundancy, so that reducing the number of parameters only reduce the learning capacity.
    On the contrary, large networks such as wide residual network with large widening factor or residual networks for ImageNet are more subject to redundant parameters.
    In that case, sharing weights makes more sense, like for CIFAR 100 where the accuracy gap is only of $0.2\%$ with parameters reduced by $26\%$.

    \begin{table}[!ht]
        \centering
        \begin{tabular}{|c|c|c|c|}
            \hline
            Dataset & Model & Param. & Acc. \\
            \hline
            \multirow{4}{*}{CIFAR 10}& ResNet-164 Share & 0.93 M & 93.8 \\
             &ResNet-92        & 0.96 M & 93.9 \\
             \cline{2-4}
             &WRN-40-4  Share  & 5.85 M & 94.9 \\
            &WRN-28-4         & 5.85 M & 95.0 \\
            \hline
            \multirow{2}{*}{CIFAR 100} &WRN-28-10 Share & 26.86 M & 79.8  \\
            & WRN-22-10       & 26.85 M & 79.55 \\
            \hline
        \end{tabular}
        \caption{Comparison of accuracies between ShaResNet and ResNets with equivalent size.}
        \label{tab:comp}
    \end{table}

    Compared to sequential networks, sharing spatial connections at stage level induces a loss of accuracy at test time.
    We now compare our ShaResNets to less deep networks with a similar number of parameters.
    Table~\ref{tab:comp} shows these results for CIFAR.
    For each shared architecture, we compare its sequential counterpart with reduced depth.
    We can draw similar conclusion as in the first paragraph.
    Sharing weights on relatively small architectures (CIFAR 10) is not more efficient than using a less deep network.
    On the contrary, Wide ResNet with a widen factor of 10 (CIFAR 100) gets a boost in accuracy using shared convolutions.
    Deeper networks benefit from mutualisation of spatial relations, the weights are better used, i.e. the ratio accuracy over network size gets better.

\begin{figure}[!ht]
        \centering
        \includegraphics[width=0.9\linewidth]{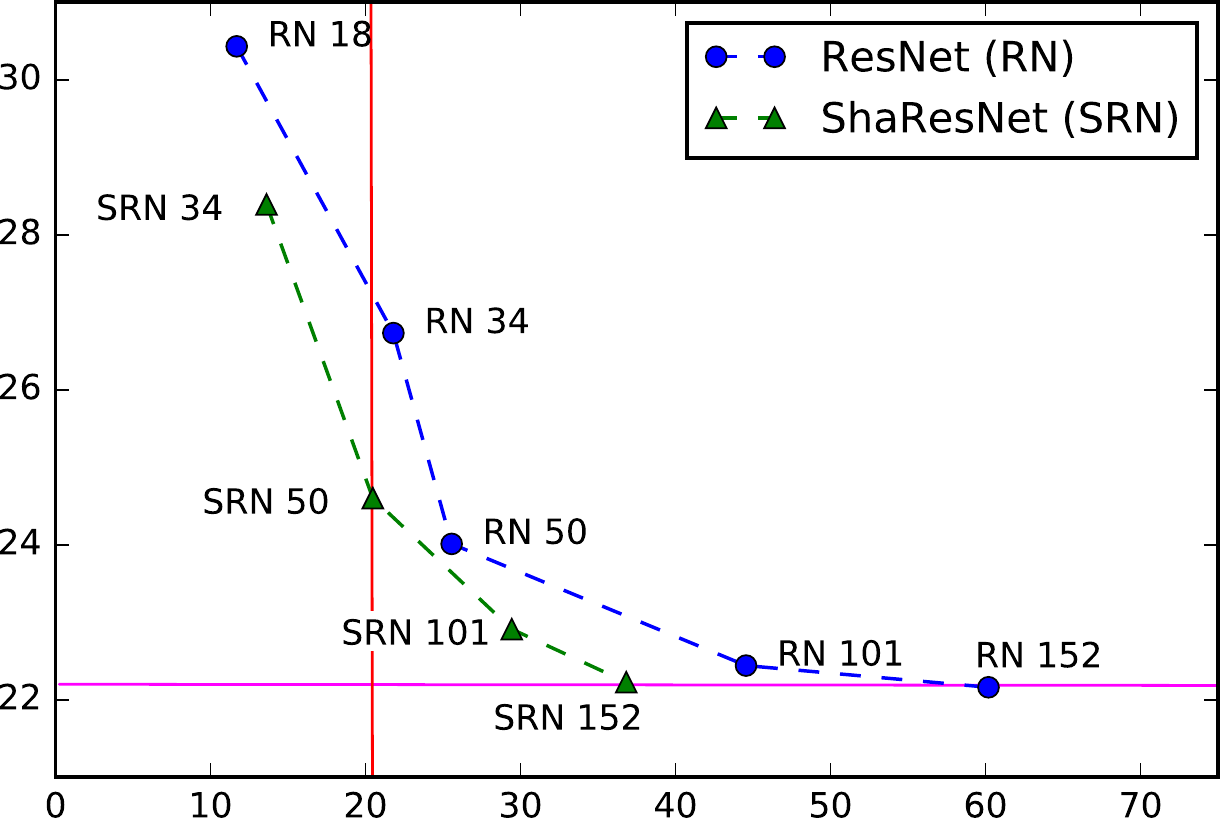}

        \textit{Red line: SRN 50 performs better than RN 34 with less parameters.}

        \textit{Magenta line: SRN 152 performs as well as RN 152 with less parameters.}

        \caption{Top 1 error (\%) of ShaResNet and ResNets function of the model size (millions of parameters).}
        \label{fig:imagenet_accu}
    \end{figure}

    On Imagenet, table~\ref{tab:acc_datasets} underlines that the sharing is more efficient as the network goes larger.
    We even reach similar accuracy (less than 0.2\% drop) for the 152 layer architecture.
    The figure~\ref{fig:imagenet_accu} presents the top-1 error function of the weight on ImageNet.
    The shared architectures plot (green curve) is situated under the blue curve (residual networks).
    It illustrates that for large networks, shared networks are more efficient than their sequential peers with similar number of parameters.

\subsection{Limitations and perspectives}

We have shown in the previous section that residual networks with shared spatial connections are particularly efficient on large networks: given a number of parameters, ShaResNets are more efficient (figure~\ref{fig:imagenet_accu}).
However, they induce a loss in terms of accuracy sometimes even leading to performances similar to networks with reduced depth (CIFAR 10 networks).
The conclusions we can draw is that, first, in relatively small networks parameters are used to their potential or at least that redundant spatial connectivities are fewer in number than for large networks.

Second, we chose an arbitrary shared structure.
We considered the 3x3 convolutions to operate similar operations for all blocks in the same stage.
This assumption may be too restrictive.
In our future work we will investigate flexible shared networks, where the sharing rate would be adaptative, function of the noise, the position in the stage and classification dataset (image size, class number).
Moreover, we would also investigate other possible splits between spatial relations and block specific operations that would require to modify the basic or bottleneck residual block structure, for example using channel wise convolution as spatial relations and 1x1 convolution for information abstraction.

%%%%%%%%%%%%%%%%%%%%%%%%%%%%%%%%%%%%%%%%%%%%%%%%%%%%%%%%%%%%%%%%%%%%%%%%%%%%
%%%%%%%%%%%%%%%%%%%%%%%%%%%%%%%%%%%%%%%%%%%%%%%%%%%%%%%%%%%%%%%%%%%%%%%%%%%%
%  ######   #######  ##    ##  ######  ##       ##     ##
% ##    ## ##     ## ###   ## ##    ## ##       ##     ##
% ##       ##     ## ####  ## ##       ##       ##     ##
% ##       ##     ## ## ## ## ##       ##       ##     ##
% ##       ##     ## ##  #### ##       ##       ##     ##
% ##    ## ##     ## ##   ### ##    ## ##       ##     ##
%  ######   #######  ##    ##  ######  ########  #######
%%%%%%%%%%%%%%%%%%%%%%%%%%%%%%%%%%%%%%%%%%%%%%%%%%%%%%%%%%%%%%%%%%%%%%%%%%%%
%%%%%%%%%%%%%%%%%%%%%%%%%%%%%%%%%%%%%%%%%%%%%%%%%%%%%%%%%%%%%%%%%%%%%%%%%%%%
\section{Conclusion}
\label{sec:conclu}

In this paper, we introduced the \textit{ShaResNet}, a new convolutional neural network architecture based on residual networks.
By sharing convolutions between residual blocks, we create neural architectures lighter than their sequential residual counterpart by $25\%$ to $45\%$ in terms of number of parameters.
The training of such network is as easy as with common residual networks.
We experimented on three classification datasets.
Shared residual architecture proved to be efficient for large networks.
We observed an accuracy gap to the corresponding residual architecture of less than $1\%$ for a substantial size reduction.
By exploiting the redundant relations of 3x3 convolutions between residual blocks, the ShaResNets make better use of the optimizable weights.
With an equivalent parameter number, we obtain better results.
We hope that these findings will help further investigation in image processing and more generally in deep learning research.

\section*{Implementation and hardware details}
The experiments uses Torch7 with neural network package.
Our code for CIFAR 10 and 100 experiments is based on the original implementation of Wide Residual Networks (\texttt{github.com/szagoruyko/wide-residual- networks}).
Similarly for Imagenet, we used the code from \texttt{github.com/facebook/fb.resnet.torch}.

Experiments were operated using Titan X (Maxwell) GPUs, one for CIFAR and two for ImageNet.

\section*{Code}

The code is available online at \url{github.com/aboulch/sharesnet}.

\section*{Acknowledgements}
This work is part of the DeLTA project at ONERA \url{delta-onera.github.io}.
This projects aims at developing innovative machine learning approaches for aerospace applications.

\footnotesize
\bibliographystyle{alpha}
\bibliography{main}

\end{document}